\documentclass[10pt,twocolumn,letterpaper]{article}

\usepackage{iccv}
\usepackage{times}
\usepackage{epsfig}
\usepackage{graphicx}
\usepackage{amsmath}
\usepackage{amssymb}

\usepackage{booktabs}
\usepackage{caption}
\usepackage{multirow}
\usepackage{color}
\usepackage{subcaption}
\usepackage{soul}
\usepackage{diagbox}
\usepackage{soul}
\usepackage{pifont}
\usepackage{makecell}
\usepackage[accsupp]{axessibility}

\usepackage[breaklinks=true,bookmarks=false]{hyperref}

\iccvfinalcopy 

\def\iccvPaperID{****} 

\ificcvfinal\fi

\newcommand\extrafootertext[1]{%
    \bgroup
    \renewcommand\thefootnote{\fnsymbol{footnote}}%
    \renewcommand\thempfootnote{\fnsymbol{mpfootnote}}%
    \footnotetext[0]{#1}%
    \egroup
 }

\begin{document}
\title{E2E-LOAD: End-to-End Long-form Online Action Detection} 

\author{
Shuqiang Cao\textsuperscript{\rm 1}\thanks{\textit{Authors contributed equally.}}
~~~~~~~~
Weixin Luo\textsuperscript{\rm 2}\footnotemark[1]~~~~~~~~
Bairui Wang\textsuperscript{\rm 2}~~~~~~~~
Wei Zhang\textsuperscript{\rm 1}\thanks{\textit{Corresponding author.}}  ~~~~~~~~
Lin Ma\textsuperscript{\rm 2}\footnotemark[2]\\
\textsuperscript{\rm 1}School of Control Science and Engineering, Shandong University \\
\textsuperscript{\rm 2}Meituan\\
{\tt\small sqiangcao@mail.sdu.edu.cn, luowx@shanghaitech.edu.cn, davidzhang@sdu.edu.cn} \\
{\tt\small \{bairuiwong, forest.linma\}@gmail.com}
}

\maketitle
\ificcvfinal\thispagestyle{empty}\fi
\begin{abstract}
Recently, feature-based methods for Online Action Detection (OAD) have been gaining traction. However, these methods are constrained by their fixed backbone design, which fails to leverage the potential benefits of a trainable backbone. 
This paper introduces an end-to-end learning network that revises these approaches, incorporating a backbone network design that improves effectiveness and efficiency.
Our proposed model utilizes a shared initial spatial model for all frames and maintains an extended sequence cache, which enables low-cost inference. We promote an asymmetric spatiotemporal model that caters to long-form and short-form modeling. Additionally, we propose an innovative and efficient inference mechanism that accelerates extensive spatiotemporal exploration.
Through comprehensive ablation studies and experiments, we validate the performance and efficiency of our proposed method. Remarkably, we achieve an end-to-end learning OAD of 17.3 (+12.6) FPS with 72.4\%~(+1.2\%), 90.3\%~(+0.7\%), and 48.1\%~(+26.0\%) mAP on THMOUS'14, TVSeries, and HDD, respectively.
The source code is available at \url{https://github.com/sqiangcao99/E2E-LOAD}.
\end{abstract}

\section{Introduction}
\label{sec:intro} 
Online Action Detection (OAD)\cite{geest2016online} has become a critical domain in computer vision, driven by its extensive applicability spanning surveillance, autonomous driving, and more.
Recent research endeavors\cite{xu2021long,chen2022gatehub,wang2021oadtr,zhao2022real} have begun embracing the Transformer architecture~\cite{vaswani2017attention} for this task. By leveraging the attention mechanism's capability for long-range interactions, these methods manifest marked improvements over their RNN-based counterparts~\cite{xu2019temporal,eun2020learning}.
Nevertheless, most existing studies~\cite{xu2019temporal,xu2021long,chen2022gatehub} rely on features from pre-trained networks. 
The dependency on a frozen backbone progressively constrains improvements in both speed and precision. Although there are efforts~\cite{yang2022colar,cheng2022stochastic} to fine-tune the backbone directly, they often fall short in balancing outstanding performance with acceptable computation costs.
\begin{figure}[t]
\centering
\scalebox{1.0}{
\includegraphics[height=6cm]{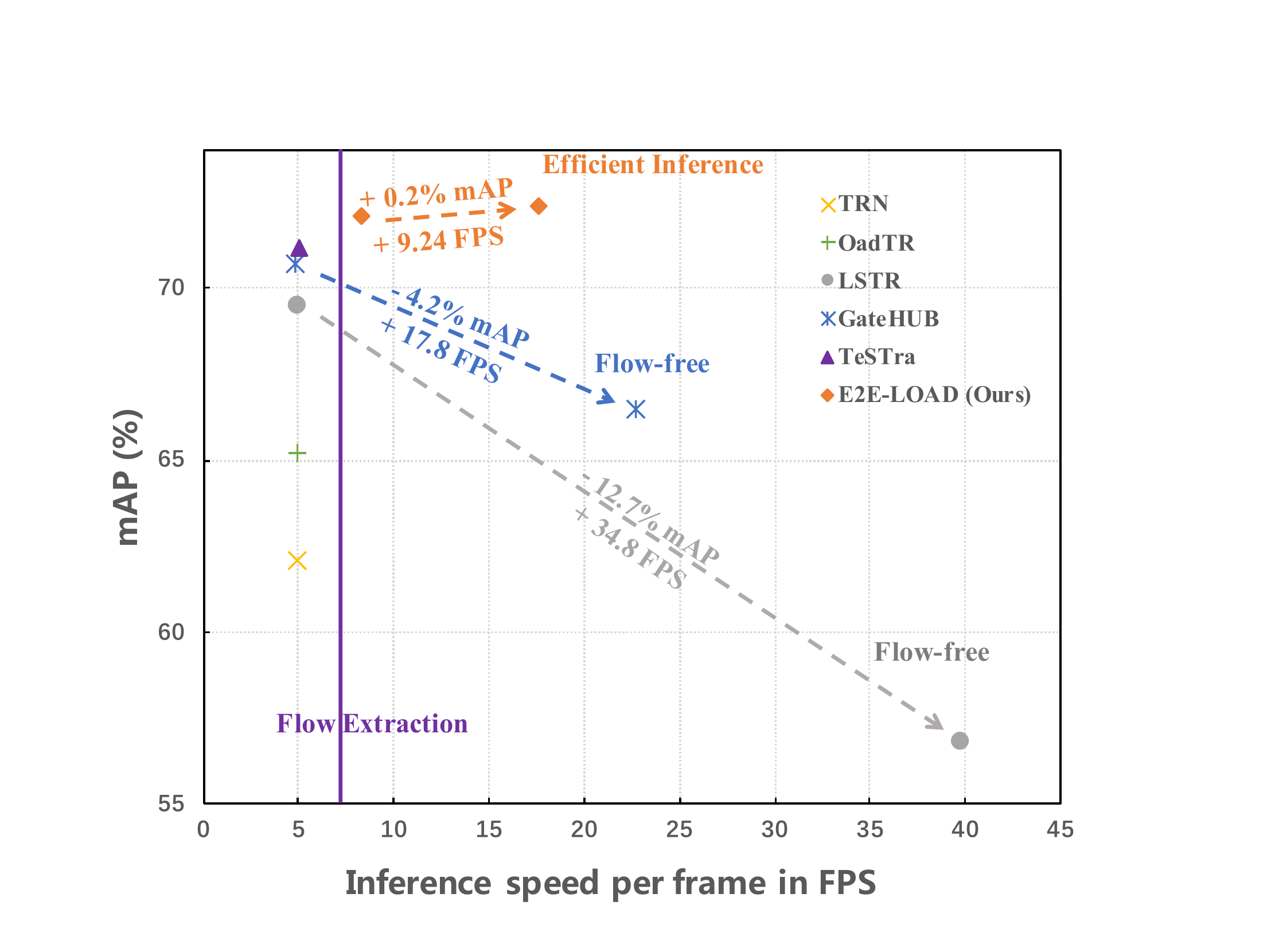}
}
\caption{\textbf{Comparison of Performance (mAP and FPS).} Methods like GateHUB~\cite{chen2022gatehub} and LSTR~\cite{xu2021long} have eliminated computation-intensive optical flow inputs to speed up inference, albeit at the cost of a considerable decline in performance. In contrast, our E2E-LOAD, benefiting from backbone design and efficient inference mechanism, achieves superior mAP and FPS.}
\label{fig:teaser}
\end{figure}
This is primarily because these feature-based methods adopt a paradigm that employs a heavy spatiotemporal backbone for individual local chunks coupled with a lightweight temporal model for chunk-wise interactions. Such an architecture often results in a less-than-ideal balance between performance and efficiency. 
Specifically, the localized employment of the heavy spatiotemporal model might not fully exploit the backbone's full potential in modeling long-term dependencies. 
Additionally, the subsequent lightweight temporal model often struggles to capture long-term relationships effectively. Moreover, this design imposes challenges for end-to-end training, as it requires the parallel execution of multiple backbone networks for feature extraction from each video chunk, resulting in substantial GPU memory consumption.
As a response, this paper proposes the design of an end-to-end learning Transformer for OAD, enhancing its scalability for practical applications.

Specifically, we introduce a novel method named the \textbf{E}nd-to-\textbf{E}nd \textbf{L}ong-form Transformer for \textbf{OAD} task, abbreviated as \textbf{E2E-LOAD}.
Our approach employs the ``Space-then-Space-time'' paradigm. Initially, raw video frames are processed by the spatial model and transformed into features, which are subsequently cached in a buffer. This technique is instrumental in managing streaming video data, as it allows for the re-utilization of these buffered features across diverse time steps, thereby significantly decreasing computational overhead.
Furthermore, the buffering mechanism boosts the model's capability to process extended historical sequences,
as it retains most frames as compact representations within the buffer, alleviating the computational burden. 
Next, we partition the sequences conserved in the cache into long-term and short-term histories and conduct spatiotemporal modeling independently. We implement a shallow branch for the long-term stream and a deep branch for the short-term stream. This asymmetric architecture promotes efficient long-form feature extraction.
Finally, we introduce a token re-usage strategy to mitigate the high computation costs of spatiotemporal interactions on extended video clips, achieving a $2\times$ speed enhancement.
Regarding implementation, we train with shorter history sequences and then increase the sequence length for inference. This technique mitigates the training expenses associated with long-term videos while enabling us to leverage the benefits of long-term context. The experiments demonstrate that this strategy effectively reduces training costs without compromising the model's effectiveness.

Through these architectural innovations and efficiency techniques, E2E-LOAD addresses the limitations inherent in feature-based methods, achieving superior effectiveness and efficiency. A comparison of E2E-LOAD with other feature-based methods is illustrated in Figure~\ref{fig:teaser}. The results underscore that our model excels in efficiency and effectiveness compared to other methods.
We perform comprehensive experiments on three public datasets: THUMOS14~\cite{idrees2017thumos}, TVSeries~\cite{geest2016online}, and HDD~\cite{ramanishka2018CVPR}. E2E-LOAD yields mAP of 72.4~(+1.2)\%, mcAP of 90.3~(+0.7)\%, and mAP of 48.1~(+26.0)\% respectively, showcasing substantial improvements. Notably, E2E-LOAD is roughly $3 \times$ faster than these methods regarding inference speed.
In summary, our key contributions are: 
($i$) We propose a unique end-to-end learning framework that integrates a stream buffer between the spatial and spatiotemporal models, thereby enhancing the effectiveness and efficiency of online data processing.
($ii$) We introduce an efficient inference mechanism that accelerates spatiotemporal attention processing through token re-usage, achieving a $2\times$ reduction in running time. 
($iii$) Our method achieves significant accuracy and inference speed advancements using only RGB frames on three public datasets, highlighting its promise for practical use in real-world scenarios.
\section{Related Works}
\label{sec:related}
{\noindent \bf Online Action Detection.} Online action detection (OAD) seeks to identify incoming frames in an untrimmed video stream instantaneously. Unlike offline video tasks, which access all the frames, only the gradually accumulated historical frames are available at each moment in OAD. 
Several methods~\cite{wang2021oadtr,eun2020learning,gao2021woad,yang2022colar} rely solely on recent video frames that span a few seconds as contextual information for the current frame. However, such approaches may overlook critical information in long-term historical frames, potentially enhancing performance. To address this, TRN~\cite{xu2019temporal} employs LSTM~\cite{hochreiter1997long} to memorize all historical information, albeit with limitations in modeling long dependencies. Recently, LSTR~\cite{xu2021long} proposed the concurrent exploration of long-term and short-term memories using Transformer~\cite{vaswani2017attention}, significantly improving action identification performance at the current frame due to the globally attended long-term history.
Beyond historical information exploration, some methods~\cite{wang2021oadtr,xu2019temporal} attempt to circumvent causal constraints by anticipating the future. OadTR~\cite{wang2021oadtr}, for instance, combines the predicted future and the current feature to identify the ongoing action. Other methods~\cite{gao2019startnet,shou2018online} concentrate on detecting the commencement of an action, with StartNet~\cite{gao2019startnet} decomposing this task into action recognition and detection of action start points. 
Recently, GateHUB~\cite{chen2022gatehub} introduced a gate mechanism to filter out redundant information and noise in historical sequences. Furthermore, Zhao \etal proposed TeSTra~\cite{zhao2022real}, a method that reuses computation from the previous step, making it highly conducive to real-time inference. Uncertaion-OAD~\cite{guouncertainty} introduces prediction uncertainty into the spatiotemporal attention for OAD.

{\noindent \bf Action Recognition.} For a comprehensive overview of classical action recognition methods, we refer the reader to the survey by Zhu \etal~\cite{zhu2020comprehensive}. Due to space constraints, we focus here on the latest works, especially those based on the Transformer paradigm~\cite{bertasius2021space,arnab2021vivit,neimark2021video,wu2022memvit,liu2022video,fan2021multiscale,li2022mvitv2}, which have achieved significant improvements in video understanding tasks. The central challenge encountered with these approaches is the substantial computational burden generated by element-wise interaction in the spatiotemporal dimension.
To address this issue, recent studies~\cite{bertasius2021space,arnab2021vivit,liu2022video} have proposed several variants of spatiotemporal attention using spatiotemporal factorization. For instance, MViT~\cite{fan2021multiscale,li2022mvitv2} introduced pooling attention to reduce the token number at different scales, while Video Swin Transformer~\cite{liu2022video} adopted the shifted window mechanism~\cite{liu2021swin} to constrain element-wise interactions within local 3D windows. Although these methods exhibit impressive spatiotemporal modeling capabilities, they are rarely tailored for OAD, where speed and accuracy are of the essence.

\begin{figure*}[t]
\centering
\includegraphics[width=0.9\textwidth]{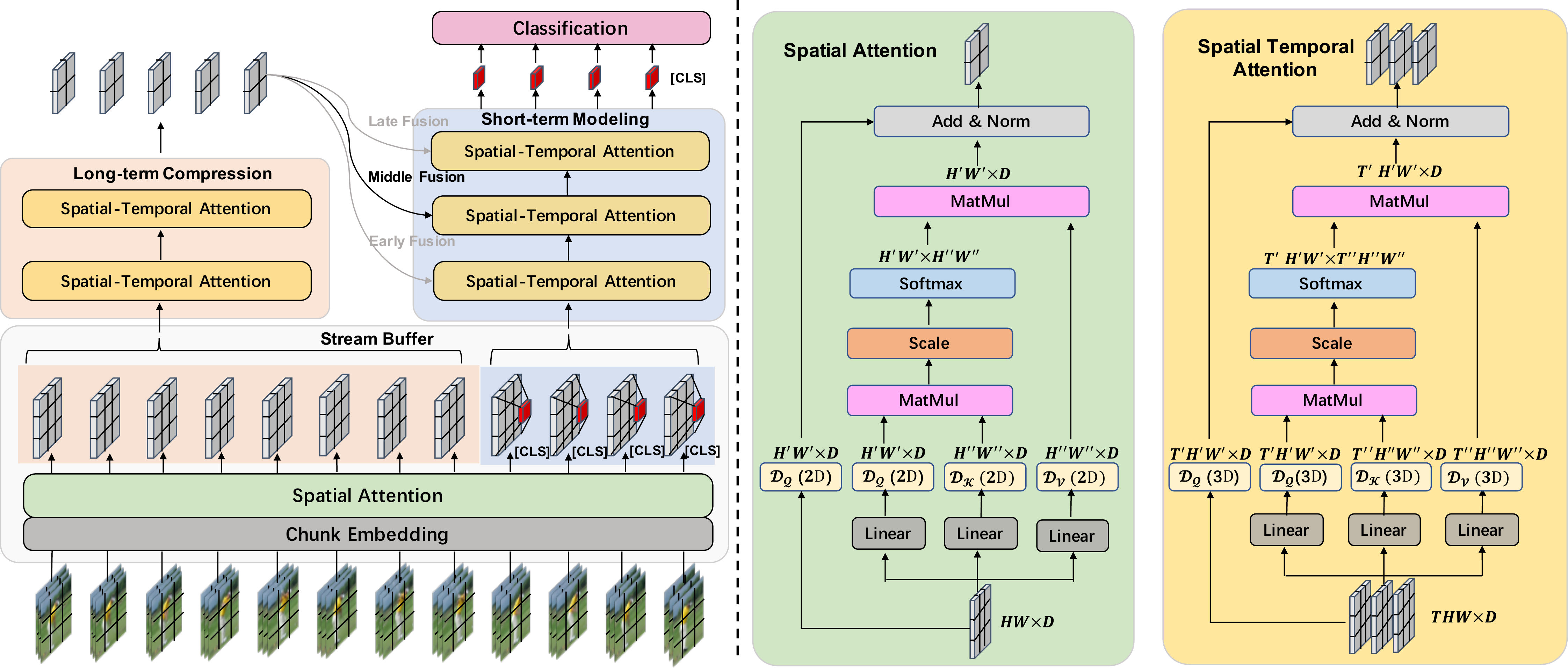}
\caption{\textbf{Overview of the Proposed E2E-LOAD.} (1) A Stream Buffer with shared chunk embedding and spatial modeling is built to reuse computed frames during inference. (2) Two asymmetric spatiotemporal modelings are designed to tackle the information with different lengths. (3) Three options are explored for the long-short-term fusion. (4) Spatial and Spatiotemporal Attention are building blocks, where $\mathcal{D}$ represents the down-sampling operation. (5) We adopt the CLS tokens to finalize the classification.} 
\label{fig:pipeline}
\end{figure*}

\section{Method}
\label{sec:method} 
\subsection{Task Definition}
Given a streaming video $\mathbf{V}=\left\{f_t\right\}_{t=-T}^0$. OAD aims to compute the action probability distribution ${\boldsymbol{y}_0} \in {[0,1]}^{C}$ for the present frame $f_0$, where $T$ indicates the count of observed frames and $C$ corresponds to the total number of action classes. Notably, ${f_1, f_2,...}$ represent future frames, which are unattainable. Unlike previous works~\cite{xu2019temporal,xu2021long,chen2022gatehub,zhao2022real} that use pre-extracted features, our E2E-LOAD directly processes raw RGB frames end-to-end. Before introducing it, we elucidate the efficient attention mechanism incorporated in our model, inspired by recent progress in video understanding~\cite{arnab2021vivit,li2022mvitv2,bertasius2021space}. 

\subsection{Efficient Attention}
Consider input sequences, $\mathbf{X}_1 \in \mathbb{R}^{N_1 \times D}$ and $\mathbf{X}_2 \in \mathbb{R}^{N_2 \times D}$, where $N_1$ and $N_2$ represent the sequence length, and $D$ signifying the channel dimension. The attention mechanism learns to assign weights to individual elements within $\mathbf{X}_2$. These elements, weighted accordingly, are then aggregated to update $\mathbf{X}_1$.
However, the complexity of this operation is positively correlated with the length of the input sequence. To address this, we employ down-sampling techniques $\mathcal{D}$ on the query (Q), key (K), and value (V) to mitigate computational complexity. 
\begin{align}
\mathbf{Q} &= \mathbf{X}_1 \mathbf{W}_q, & \mathbf{\hat{Q}} &= \mathcal{D}(\mathbf{Q}) \\
\mathbf{K} &= \mathbf{X}_2 \mathbf{W}_k, & \mathbf{\hat{K}} &= \mathcal{D}(\mathbf{K}) \\
\mathbf{V} &= \mathbf{X}_2 \mathbf{W}_v, & \mathbf{\hat{V}} &= \mathcal{D}(\mathbf{V})
\end{align}
Due to its empirically superior performance, we adopt convolution with strides for down-sampling. This technique has been widely used in action recognition tasks and facilitates spatiotemporal attention acceleration~\cite{liu2021swin,li2022mvitv2,fan2021multiscale}.
Next, we apply the attention operation on these tensors to generate the down-sampled feature map, $\hat{\mathbf{X}}_1 \in \mathbb{R}^{N^{\prime} \times D^{\prime}}$. To align their sequence length, a residual connection from $\mathbf{X}_1$ to $\hat{\mathbf{X}}_1$ is utilized along with a pooling operation $\mathcal{D}$. The resulting sequence $\Tilde{\mathbf{X}}_1 \in \mathbb{R}^{N^{\prime} \times D^{\prime}}$ is then processed by MLP to produce the final output. We define the attention mechanism as follows, excluding layer normalization for simplicity:
\begin{align}
\hat{\mathbf{X}}_1 &= \operatorname{Softmax}\left(\mathbf{\hat{Q}}{\mathcal{\mathbf{K}}}^T / \sqrt{\mathbf{D}'}\right) {\mathcal{\mathbf{V}}} \\
\widetilde{\mathbf{X}}_1 &= \hat{\mathbf{X}}_1 + \mathcal{D}(\mathbf{X}_1) \\
\operatorname{Attn}(\mathbf{X}_1,\mathbf{X}_2) &= \operatorname{MLP}(\widetilde{\mathbf{X}}_1)
\end{align}

\subsection{Architecture}
We present the E2E-LOAD architecture, which employs a Stream Buffer (SB) to extract and cache the spatial representations of incoming frames. These representations are then divided into two parts. The older, longer part is directed to a Long-term Compression (LC) branch to compress temporal resolution. In contrast, the newer, shorter piece is sent to a Short-term Modeling (SM) branch to model the recent context carefully. Finally, these two representations are fused via a Long-Short-term Fusion (LSF) module to predict the latest frame. During inference, we introduce an Efficient Inference (EI) technique to accelerate the spatiotemporal exploration of SM. Figure~\ref{fig:pipeline} depicts the structure of E2E-LOAD. The details of each module are discussed in the sections that follow.

\subsubsection{Chunk Embedding}
Commonly in offline video recognition~\cite{arnab2021vivit,bertasius2021space}, 2D or 3D patches are uniformly sampled from videos and projected into a token sequence for the Transformer encoder. 
For OAD, previous approaches~\cite{xu2021long,zhao2022real,chen2022gatehub} identify ongoing actions at the chunk level, where each chunk consists of several consecutive video frames. 
Following this configuration, we evenly sample $t$ frames from each chunk of $\tau \times H \times W$, partitioning it into $n_h \cdot n_w$ 3D patches of $t \times h \times w$ along the spatial dimension, where $n_h=\left\lfloor\frac{H}{h}\right\rfloor, n_w=\left\lfloor\frac{W}{w}\right\rfloor$. 
The resulting 3D patches are then projected to chunk embedding $\mathbf{E}_t \in \mathbb{R}^{(n_h \cdot n_w) \times D}$ using Chunk Embedding (CE). This embedding process allows each token to incorporate local spatiotemporal clues, which is beneficial for fine-grained recognition. 
It's noteworthy that feature-based methods~\cite{xu2021long,chen2022gatehub,zhao2022real} typically rely on heavy spatiotemporal backbones, such as two-stream~\cite{simonyan2014two} and TimeSFormer~\cite{bertasius2021space}, to extract chunk features, while often employing lightweight modules for chunk interaction to maintain overall efficiency. This inflexible and unbalanced design hinders improvements in OAD's efficiency and effectiveness.
\subsubsection{Stream Buffer}
In online scenarios, OAD models receive one frame at a time, using existing memory to identify ongoing actions. However, most action recognition models necessitate temporal interaction among these frames, introducing inefficiencies in processing online videos because such design hinders the reuse of intermediate frame representations due to the sequence evolving over time.
To tackle this challenge, we introduce a spatial attention module with a buffer for storing intermediate per-frame features. The spatial attention projects the raw frames into compact yet semantically rich representations, which can be reused at different time steps, alleviating the burden of subsequent spatiotemporal modeling. The cached memories, $\mathbf{M}_t$, are expressed as follows:
\begin{align}
\mathbf{\hat{E}}_t &= \operatorname{Attn_{S}} (\mathbf{E}_t) \\
\mathbf{M}_t &= [\mathbf{\hat{E}}_{t-T+1}, \dots, \mathbf{\hat{E}}_{t-1}, \mathbf{\hat{E}}_t]
\end{align}
where $T$ represents the memory length and $t$ indexes the timestamps. $T$ can be large, which is crucial for long-term understanding. 
Lastly, we append [CLS] tokens to each chunk for action classification, as OAD requires fine-grained, chunk-level predictions. 
\subsubsection{Short-term Modeling}
\label{sec:method:sm}
Previous works~\cite{xu2019temporal,xu2021long} often used heavy spatiotemporal backbones~\cite{simonyan2014two,bertasius2021space} to extract features from chunks. These features were then pooled into a 1D format, with lightweight temporal modeling~\cite{hochreiter1997long,vaswani2017attention} creating dependencies. While this approach provided some computational efficiency, it overlooked the importance of spatiotemporal modeling among chunks for fine-grained action classification.
In contrast, our method employs the Stream Buffer (SB) module to capture spatial features within each chunk. We then apply spatiotemporal modeling among these chunks. Such a ``Space-then-Space-time'' design fully uses the backbone's representational capacity for long-range dependencies, without wasting excessive resources on computing dependencies within individual chunks. Consequently, our end-to-end trained framework delivers improvements in both efficiency and effectiveness.
Concretely, we take the $T_{S}$ most recent chunks $\mathbf{M}^{S}_t=\left[\hat{\mathbf{E}}_{t-T_{S}+1}, ..., \hat{\mathbf{E}}_{t-1}, \hat{\mathbf{E}}_t\right]$ from the Stream Buffer as the input of Short-term Modeling (SM). Then, the stacked multi-layer attentions build the spatiotemporal interactions among the inputs with $T_S \cdot n_h \cdot n_w$ tokens. 
\begin{align}
\hat{\mathbf{M}}^{S}_{t} &= \operatorname{Attn_{ST}}(\mathbf{M}^{S}_{t})
\end{align}
Additionally, we employ a causal mask to the short-term history to block any future interactions for each token, in line with previous works~\cite{xu2021long}. Upon completing the spatiotemporal exploration of the current context, we feed the [CLS] token of the last frame to the classifier for the action prediction.
\subsubsection{Long-term Compression}
Over time, extensive video frames are cached within the streaming buffer. These frames may contain critical information that can assist in identifying the current frame.
Therefore, we compress the long-term historical sequences into several spatiotemporal feature maps, providing an extended time-scale context for the short-term modeling (SM) module.
Specifically, we sample the long-term history from $\mathbf{M}_t$, where $\mathbf{M}^{L}_{t}=\left[\hat{\mathbf{E}}_{t-T_S-T_L+1}, \ldots, \hat{\mathbf{E}}_{t-T_S-1}, \hat{\mathbf{E}}_{t-T_S}\right]$, and $T_L$ represents the length of the long-term memory.
Then we utilize spatiotemporal attention to compress $\mathbf{M}_t^{L}$ using a larger down-sampling rate than that used in SM.
To achieve efficiency, we construct a shallow compression module with $L_{LC}$ attention layers since the correlation between long-term history and current actions is generally weaker compared to short-term history.
This module progressively reduces the spatial and temporal resolution. Through several stages, the resulting tokens aggregate the most critical spatiotemporal clues. 
More importantly, before ${\mathbf{M}}^{L}_{t}$ is fed to the compression module, we detach $\mathbf{M}_t^L$ to stop back-propagation from $\mathbf{M}_t^L$ to the Stream Buffer. This step is taken as we empirically observe a training frustration if such gradient truncation is not applied. It is worth noting that the Short-term Modeling has already provided gradients for training the Stream Buffer. The ``stop gradient'' operator, used to implement this detachment, is denoted as $\operatorname{sg}(\cdot)$. The formulation of this process is illustrated below.
\begin{align}
{\mathbf{M}}^{L}_{t} &= \operatorname{sg}([(\hat{\mathbf{E}}_{t-T_L+1}), \ldots, (\hat{\mathbf{E}}_{t-T_S-1}), (\hat{\mathbf{E}}_{t-T_S})]) \\
\hat{\mathbf{M}}_{t}^{L} &= \operatorname{Attn_{ST}}({\mathbf{M}}_t^{L})
\end{align}
where $\hat{\mathbf{M}}_{t}^{L} \in \mathbb{R}^{T_L^{\prime} \cdot n_h^{\prime} \cdot n_w^{\prime} \times D}$ and  $T_L^{\prime},n_h^{\prime}$ and $n_w^{\prime}$ represent the resolution of the compressed historical representations.  
\subsubsection{Long-Short-term Fusion} 
The fusion of long-term and short-term histories is a critical technical aspect that significantly impacts the ability of each branch to learn better representations of their characteristics. Therefore, we explore various fusion operators and positions to achieve more effective integration between the long-term compression$\hat{\mathbf{M}}_t^{L}$ and the short-term histories $\hat{\mathbf{M}}_t^{S}$. 
Unlike previous work~\cite{simonyan2014two,feichtenhofer2019slowfast,wu2022memvit,xu2021long}, we aim to fuse them in space-time. This approach allows $\hat{\mathbf{M}}_t^{S}$ to discover and accumulate $\hat{\mathbf{M}}_t^{L}$ through more fine-grained spatiotemporal cubes rather than relying on whole image representations. The details of this method are as follows.\\
\noindent\textbf{Fusion Operation.}~For the \emph{cross-attention~(CA)} based fusion, we take the compressed long-term history $\hat{\mathbf{M}}_t^{L}$ as the key and value tokens, and the short-term trend $\hat{\mathbf{M}}_t^{S}$ as the query tokens, to perform cross-attention. In contrast, we reuse the spatiotemporal attention in short-term modeling for the \emph{self-attention~(SA)} based fusion. This is done by concatenating $\hat{\mathbf{M}}_t^{L}$ with $\hat{\mathbf{M}}_t^{S}$ as its key and value tokens. While this approach does not introduce extra parameters, it increases computational costs. \\ 
\noindent\textbf{Fusion Position.}~One intuitive method, referred to as \emph{Late Fusion}, is to perform fusion after the long-term and short-term memories have been fully explored, similar to previous OAD approaches~\cite{xu2021long,chen2022gatehub,zhao2022real}. 
In contrast, \emph{Early Fusion} integrates the compressed long-term history with the intermediate representations within a layer of the Short-term Modeling module, allowing the subsequent layers to explore the fused representations further.
\subsection{Efficient Inference}
Although the proposed Stream Buffer can reuse the computed features and accelerate the online inference, we observe a significant consumption of inference time for spatiotemporal exploration in Short-term Modeling~(SM). To address this, we propose Efficient Inference (EI) to accelerate SM. At each step, Regular Inference~(RI) requires updating all the frames within the short-term window. The EI directly reuses the results of the $T_S-1$ frames from the previous moment. 
As such, only the feature of the single latest frame needs to be calculated via cross-attention, with the computational complexity being reduced from $\mathcal{O}(T_{S}^{2})$ to $\mathcal{O}(T_{S})$. Specifically, EI is formulated as follows, where $\mathbf{X}_{[1:T_S]}^{t}$ and $\mathbf{Y}_{[1:T_S]}^{t}$ is the input and output of the spatiotemporal attention at time $t$, respectively:
\begin{align}
\mathbf{Y}_{T_S}^t &=\operatorname{Attn_{ST}}\left(\mathbf{X}_{T_S}, \mathbf{X}_{\left[1: T_S\right]}\right) \\
\mathbf{Y}_{\left[1: T_S\right]}^t  &=\operatorname{Concatenate}\left(\mathbf{Y}_{\left[2: T_S\right]}^{t-1}, \mathbf{Y}_{T_S}^t\right) 
\end{align}
For RI, the receptive field is fixed with $T_S$, and causal self-attention updates all the frames in the window. Instead, the proposed EI reuses the computed features of the $T_S-1$ overlapped frames that contain all the information in the window from the last moment. So the receptive field becomes recurrent and expands from the beginning to the current moment, introducing long-term context to short-term history as a complement to LC.
Moreover, the EI mechanism does not modify the training process. While this introduces differences between training and testing, the token-reuse strategy employed during testing does not result in information loss as we use a causal mask to cut off the token's connection to the future during training. Instead, this strategy allows us to gain information outside the window and is efficient for long video understanding. 
\begin{table*}[t]
\scriptsize
\centering
\begin{tabular}{lccccccc}
\toprule
\multirow{2}{*}{Method} & \multicolumn{3}{c}{Architecture} & \multirow{2}{*}{THUMOS'14 / mAP (\%)} & \multirow{2}{*}{TVSeries / mcAP (\%)} & \multirow{2}{*}{HDD / mAP (\%)} & \multirow{2}{*}{FPS} \\
\cmidrule{2-4}& CNN & RNN & Transformer & & & & \\ 
\midrule
TRN~\cite{xu2019temporal} & \checkmark &\checkmark & & $62.1$ & $86.2$ & $29.2^{\star}$ & $4.99$ \\ 
IDN~\cite{eun2020learning} & \checkmark &\checkmark & & $60.3$ & $86.1$ & - & -\\ 
FATS~\cite{YoungHwiKim2021TemporallySO}& \checkmark &\checkmark & & $59.0$ & $84.6$ &- & -\\ 
PKD~\cite{zhao2020privileged}&\checkmark & & & $64.5$ & $86.4$ &- & -\\ 
WOAD~\cite{gao2021woad} & \checkmark&\checkmark & & $67.1$  & - &- & - \\ 
OadTR~\cite{wang2021oadtr} &\checkmark & &\checkmark & $65.2$ & $87.2$ & $29.8$ & $4.97$\\ 
Colar~\cite{yang2022colar} &\checkmark & & \checkmark & $66.9$ & $88.1$ & $30.6$ & -\\ 
LSTR~\cite{xu2021long} &\checkmark & &\checkmark & $69.5$ & $89.1$ & -& $4.92$\\
GateHUB~\cite{chen2022gatehub} &\checkmark & &\checkmark & $70.7$ & $89.6$ & $32.1$ & $4.85$ \\
Uncertain-OAD~\cite{guouncertainty}&\checkmark & &\checkmark & $69.9$ & $89.3$ & $30.1$ & $5.03$ \\
TeSTra~\cite{zhao2022real}&\checkmark & &\checkmark & $71.2$ & - & - & -\\
\midrule
E2E-LOAD &  & &\checkmark & $\mathbf{72.4}$ & $\mathbf{90.3}$ & $\mathbf{48.1^{\star}}$ & $\mathbf{17.30}$\\
\bottomrule
\end{tabular} 
\caption{\textbf{Performance Comparison with Different Methods on THUMOS'14, TVSeries, and HDD.} For THUMOS'14 and TVSeries, the evaluated methods utilize features pre-trained on Kinetics or ActivityNet as input. 
For HDD, results marked by $\star$ indicate RGB data is used as input. Otherwise, sensor data is used as input.
The mAP is reported for THUMOS'14 and HDD, while the mcAP is reported for TVSeries. The FPS column represents the inference speed, including the time taken for feature extraction. The architectures of the compared models, including Convolution, RNN, and Transformer, are also provided for a comprehensive comparison.}
\label{tab:sota}
\end {table*}
\subsection{Objective Function}
Following LSTR~\cite{xu2021long} and GateHUB~\cite{chen2022gatehub}, we apply a cross-entropy loss over all the short-term frames, given by:
\begin{equation}
\mathcal{L}=-\sum_{i=t-T_S+1}^{t} \sum_{j=1}^{C} \boldsymbol{y}_{i, j} \log \hat{\boldsymbol{y}}_{i, j}
\end{equation}
where $y_{i}$ represents the ground truth for the $i^{th}$ frame, and $\hat{y_{i}}$ corresponds to the predicted probabilities across $C$ classes.
\section{Experiments}
\label{sec:exp}
We evaluate all the methods on the following datasets: {THUMOS'14}~\cite{idrees2017thumos}, {TVSeries}~\cite{geest2016online}, and {HDD}~\cite{ramanishka2018CVPR}. Please refer to the supplementary material for detailed information about the dataset introduction, hyperparameter settings, training procedures, and evaluation metrics.
\begin{table*}[t]
    \begin{subtable}[t]{0.32\linewidth}
       \centering
        \small
        \caption{\label{tab:exp:fusion_pos}}
            \scalebox{0.7}
            {
            \begin{tabular}{ccc|cc}
            \toprule
            Baseline & LC+LSF & \begin{tabular}[c]{@{}c@{}}
            EI
            \end{tabular}  & mAP $(\%)$ & FPS\\
            \midrule
            $\checkmark$ &  &  & $71.2$ & $9.1$ \\
            $\checkmark$ & $\checkmark$ &  & $72.2$ & $8.7$ \\
            $\checkmark$ & & $\checkmark$ & $71.5$ &  $\mathbf{19.5}$ \\ 
            $\checkmark$ & \checkmark & $\checkmark$ & $\mathbf{72.4}$ & $17.3$\\
            \bottomrule
            \label{tab:ablation}
            \end{tabular}
            }
    \end{subtable}
    \begin{subtable}[t]{0.26\linewidth}
          \centering
            \small
            \caption{\label{tab:exp:lc}}
            \scalebox{0.7}{
            \begin{tabular}{c|ccc} 
                \toprule
              Compression Factor &  mAP $(\%)$ & FPS \\
                \midrule
                $\times 4, \times 2, \times$ 1 & $70.8$ & $18.9$ \\
                $\times 4, \times 1, \times$ 1 & $70.6$ & $18.7$ \\
                $\times 2, \times 2, \times$ 1 & $71.8$ & $18.7$ \\
                $\times 2, \times 2, \times 1, \times$ 1 & $72.4$ & $17.3$ \\
                \bottomrule
            \end{tabular}
            }
    \end{subtable}
    \begin{subtable}[t]{0.2\linewidth}
       \centering
        \small
        \caption{\label{tab:exp:fusion_pos}}
        \scalebox{0.7}{
        \begin{tabular}{cc} 
            \toprule
            Layer Index &  mAP $(\%)$ \\
            \midrule
            $1~(early)$ & $70.5$ \\
            $5~(middle)$ & ${72.4}$ \\
            $7~(middle)$ & $71.7$ \\
            $11~(late)$ & $71.5$ \\
            \bottomrule
        \end{tabular}
        }
    \end{subtable}
    \begin{subtable}[t]{0.2\linewidth}
      \centering
        \small
        \caption{\label{tab:exp:fusion_ops}}
        \scalebox{0.7}{
            \begin{tabular}{@{}l|cccc@{}}
            \toprule
            Variants & \begin{tabular}[c]{@{}l@{}}
            mAP  (\%)  \end{tabular}  
            &  FPS \\
            \midrule
            CA@5 & $72.4$  & $17.3$ \\
            \midrule
            SA@5 & $70.3$  & $17.5$ \\
            SA@7 & $70.8$  & $17.5$ \\
            \bottomrule
            \end{tabular}
        }
    \end{subtable}
    \caption{\textbf{Ablation Studies} (a)~Impact of the proposed components, \ie LC+LSF and EI. (b)~Design choice of temporal downsample rate at each layer for LC module.~(c)~Design choice of the position to perform the fusion. (d)~Design choice of the fusion operators.} 
\end{table*}
\subsection{Comparison of the State-of-the-art Methods.}
As illustrated in Table~\ref{tab:sota}, we compare our proposed E2E-LOAD method with existing approaches on THUMOS'14~\cite{idrees2017thumos}, TVSeries~\cite{geest2016online}, and HDD~\cite{ramanishka2018CVPR} to validate the effectiveness of our model. 
These methods encompass architectures such as CNN~\cite{shou2017cdc}, RNN~\cite{xu2019temporal,qu2020lap}, and Transformer~\cite{xu2021long,chen2022gatehub}. For TVSeries and THUMOS'14, previous works~\cite{shou2017cdc,xu2019temporal,xu2021long,chen2022gatehub} utilize RGB and flow features with two-stream models~\cite{simonyan2014two}. In the case of HDD, TRN~\cite{xu2019temporal} uses RGB and sensor data, while GateHUB~\cite{chen2022gatehub} and OadTR~\cite{wang2021oadtr} rely solely on sensor data. Our experiments only employ the RGB modality as input.
As evident from Table~\ref{tab:sota}, E2E-LOAD outperforms all existing methods in terms of both effectiveness and efficiency across the three benchmark datasets. E2E-LOAD achieves a mcAP of 90.3~(+0.7)\% on TVSeries, becoming the first method to surpass 90\% on this dataset. The complexity of the TV series context underscores the critical role of spatiotemporal attention, thus validating the effectiveness of our proposed approach. Furthermore, E2E-LOAD reaches remarkable performances of 48.1~(+26.0)\% and 72.4~(+1.2)\% on HDD and THUMOS'14, respectively. Additionally, E2E-LOAD achieves an inference speed of 17.3 FPS, making it $3\times$ faster than all existing methods, requiring both RGB and optical flow inputs.
\begin{figure*}[t]
\centering
\subcaptionbox{\label{fig:exp:sm_depth}}{\includegraphics[width=4cm]{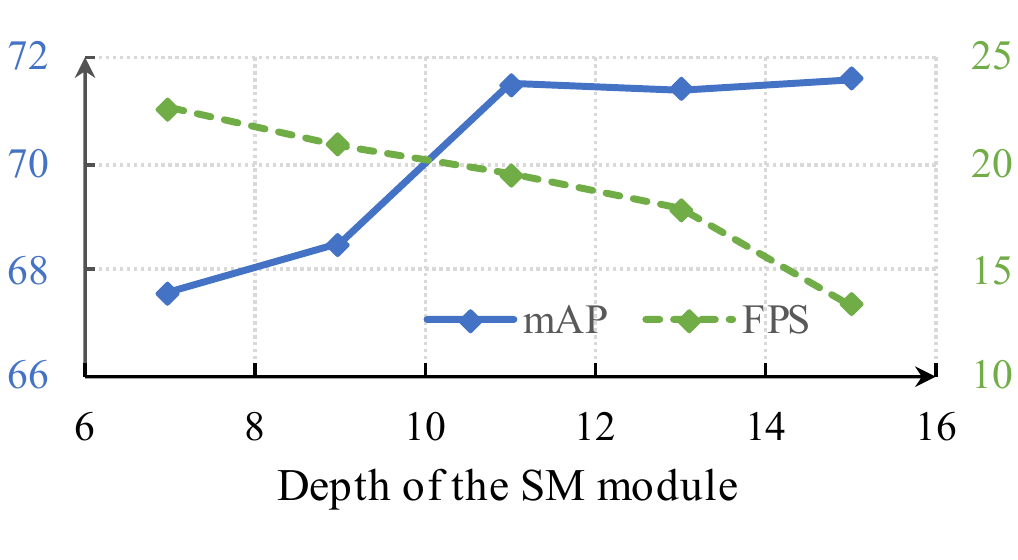}} 
\subcaptionbox{\label{fig:exp:long_len}}{\includegraphics[width=4cm]{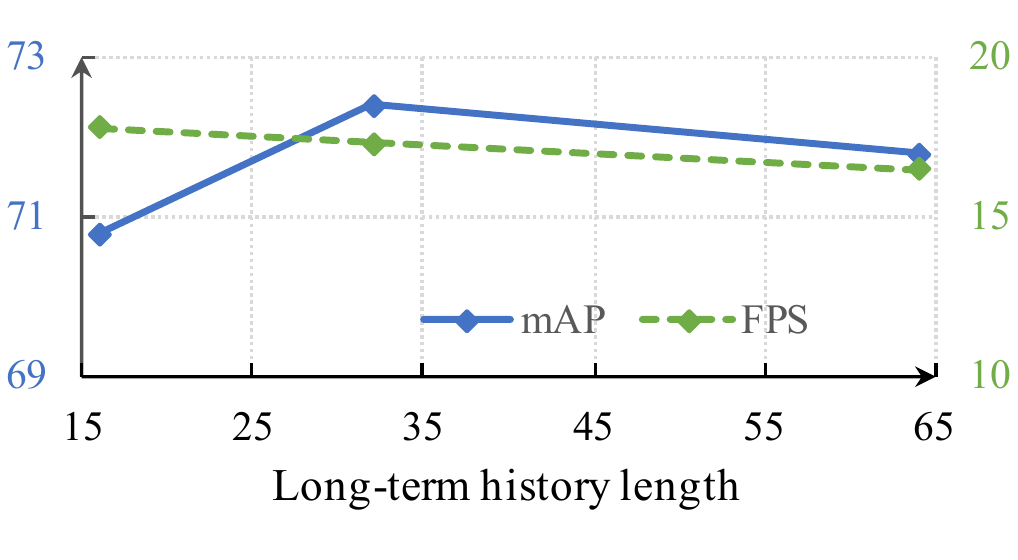} } 
\subcaptionbox{\label{fig:exp:short_len}}{\includegraphics[width=4cm]{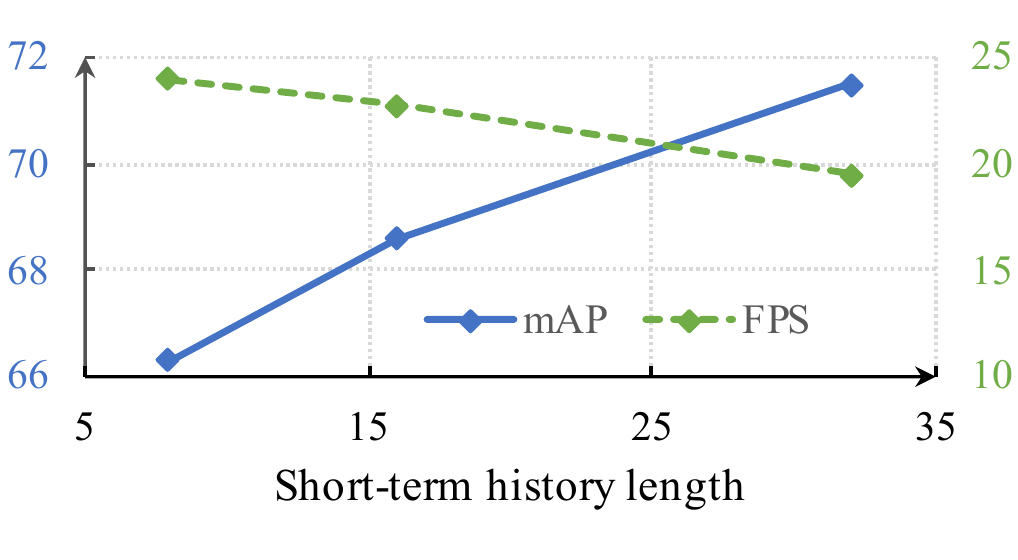} } 
\subcaptionbox{\label{fig:generalization}}{\includegraphics[width=4cm]{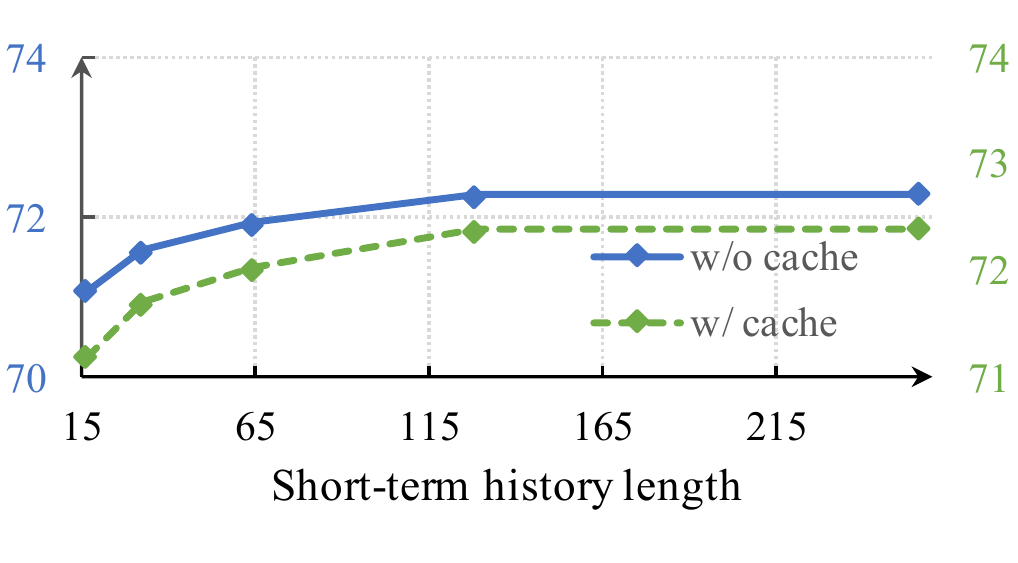} }
\caption{\textbf{Ablation Studies.} (a)~The trade-off of SM module with different depth. (b)~The trade-off of LC module with different lengths of long-term history (c)~The trade-off of SM module with different lengths of short-term history. (d)~The long sequence generalization for LC is trained with 32 frames and tested with different lengths.}
\end{figure*}
\subsection{Ablation Study}
In this section, we conduct ablation experiments to assess each component of the E2E-LOAD model. Unless explicitly stated otherwise, all experiments were performed on the THUMOS'14 dataset, with an evaluation conducted using a history length of $T_L=128$. 
\subsubsection{Impact of Each Component}
We design different configurations of the proposed E2E-LOAD as follows. The performance, in terms of FPS and mAP, is reported in Table~\ref{tab:ablation}. \\ 
\noindent\textbf{Baseline.} The \emph{Baseline} configuration only considers short-term historical frames as input. It includes the Stream Buffer (SB) and Short-term Modeling (SM) modules. The SB module caches incoming chunks as feature maps via spatial attention. Subsequently, the SM module aggregates the short-term spatial features for spatiotemporal modeling. The resulting [CLS] tokens of each chunk are then fed to a fully connected layer for classification. \\
\noindent\textbf{Baseline+LC+LSF.} 
To incorporate long-term history into the \emph{Baseline}, we introduce the Long-term Compression~(LC) module to generate compact representations of long-term history. The Long-Short-term Fusion~(LSF) then integrates the spatiotemporal cues from this long period into the short-term memory, aiding in identifying ongoing actions. \\
\noindent\textbf{Baseline+EI.} 
Our proposed Efficient Inference (EI) technique significantly accelerates the spatiotemporal attention in the SM module. We apply this technique to the \emph{Baseline} model to validate its efficiency. \\ 
\noindent\textbf{Baseline+LC+LSF+EI~(E2E-LOAD).} This configuration combines LC, LSF, and EI with the \emph{Baseline} to form the proposed E2E-LOAD model. It leverages informative long-term historical tokens while ensuring robust inference efficiency.
\begin{table}[t]
    \centering
    \footnotesize
    \scalebox{0.9}{
    \begin{tabular}{lcccc}
        \toprule
        \multirow{2}{*}{Method} & \multirow{2}{*}{Training} & \multicolumn{2}{c}{Architecture} & \multirow{2}{*}{mAP~(\%)} \\
        & & \multicolumn{1}{c}{RGB} & \multicolumn{1}{c}{Flow} & \\
        \midrule
        \multirow{6}{*}{LSTR~\cite{xu2021long}} & Feat. & TSN~\cite{wang2016temporal} & - & $56.8$ \\ 
        & E2E & TSN & - & $59.2$ \\
        & Feat. & MViT~\cite{fan2021multiscale} & - & $60.7$ \\
        & Feat. & TSN & TSN & $69.5$ \\
        & Feat. & TimSformer~\cite{bertasius2021space} & TSN & $69.6$  \\
        & Feat. & MViT & TSN & $71.2$ \\
        \midrule
        \multirow{2}{*}{GateHUB~\cite{chen2022gatehub}} & Feat. & TSN & TSN & $70.7$ \\ 
        & Feat. & TimeSformer & TSN & $72.5$ \\ 
        \midrule
        \multirow{2}{*}{TeSTra~\cite{zhao2022real}} & Feat. & TSN & TSN & $71.2$ \\ 
        & Feat. & MViT & TSN & $71.6$ \\
        \midrule
        \multirow{1}{*}{E2E-LOAD~\cite{chen2022gatehub}} & E2E & MViT & - & $72.4$ \\ 
        \bottomrule
    \end{tabular}
    }
    \caption{\textbf{Comparison of Performance with Recent Methods Under Different Configurations}. ``Feat.'' refers to training with a fixed backbone, while ``E2E'' signifies end-to-end training.}
    \label{tab:exp:fea}
\end{table}
As illustrated in Table~\ref{tab:ablation}, the \emph{Baseline} attains mAP of 71.2\% with only RGB frames, which is competitive to the state-of-the-art approach~\cite{zhao2022real}, underscoring the potential of spatiotemporal module for long-term modeling. 
Moreover, leveraging long-term context, the \emph{Baseline+LC+LSF} surpasses the \emph{Baseline} by over 1.0\%. In addition, the \emph{Baseline+EI} configuration achieves a 10.4 FPS improvement and a 0.3\% enhancement in mAP. We attribute this improvement to the reuse of tokens, which may preserve valuable long-term historical information. By combining these techniques, \emph{Baseline+LC+LSF+EI~(E2E-LOAD)} stands out by delivering the best performance compared to the other variants.
\subsubsection{Analysis of the Backbone Design.} 
\label{sec:exp:arch}
The previously featured-based approaches~\cite{xu2019temporal,xu2021long,chen2022gatehub,zhao2022real} employed a two-stream TSN~\cite{wang2016temporal} for feature extraction, whereas the proposed E2E-LOAD relies on a Transformer architecture. To isolate the effects of different architectures on performance and underscore the value of the proposed framework, we utilize advanced Transformer-based video backbones~\cite{bertasius2021space,arnab2021vivit} for training existing methods~\cite{xu2021long,zhao2022real,chen2022gatehub}. In alignment with previous work~\cite{xu2021long,chen2022gatehub}, we take each chunk as input to the backbone network and treat the [CLS] token as the representative feature of that chunk.
As illustrated in Table~\ref{tab:exp:fea}, the switch from TSN to MViT as the model's backbone led to a significant performance increase for LSTR, from 56.8\% to 60.7\%. However, this performance still falls short of the two-stream model, even though MViT is a state-of-the-art spatiotemporal backbone. Incorporating optical flow input further enhanced its performance, from 69.5\% to 71.2\%. This emphasizes the strong dependence of feature-based approaches on optical flow, a conclusion also reached by methods~\cite{chen2022gatehub, zhao2022real}. Such dependency stems from the inherent constraints of the prior feature-based framework, which applies a spatiotemporal backbone to each local chunk, thereby limiting its ability to capture long-term dependencies. Therefore, optical flow is required to augment motion information.
In contrast, our E2E-LOAD integrates lightweight spatial attention for each chunk and spatiotemporal attention across different chunks. This design enables the comprehensive utilization of long-term dependencies through the Transformer by end-to-end training. Consequently, we observed an improvement in performance from 71.2\% to 72.4\%. Furthermore, as depicted in Figure~\ref{fig:teaser}, our approach overcomes the need for optical flow, yielding substantial improvements in inference speed (+4 FPS).
\subsubsection{Analysis of the Training Cost.}
Previous studies~\cite{xu2019temporal,xu2021long,zhao2022real} typically leveraged a two-stream network~\cite{simonyan2014two} for feature extraction, with subsequent model training based on these derived features. In contrast, our approach involved the end-to-end training of the entire model. We conducted a comparative analysis of the end-to-end training costs between LSTR~\cite{xu2021long} and our method. Here both models solely utilize RGB frames, and the batch size is 16.
From Table~\ref{tab:training_cost}, we can observe that LSTR's memory consumption for end-to-end training is substantial, even when utilizing only the RGB branch of the two-stream network. In contrast, E2E-LOAD demonstrates marked improvements in several key areas when end-to-end training (E2E (L+S)) is employed: it boosts mAP by 13.2\%, reduces memory consumption by $8 \times 14.5$GB, and decreases the number of parameters by 52.4M. These enhancements stem from our framework's novel integration of the Stream Buffer and Short-term Modeling. The Stream Buffer efficiently mitigates the costs associated with processing extensive frames, while the Short-term Modeling adeptly captures long-term dependencies through spatiotemporal modeling. E2E-LOAD's ability to achieve end-to-end training with fewer resources while outperforming previous methods accentuates the superior efficacy of the proposed framework.
\begin{table}[t]
    \centering
    \footnotesize
    \scalebox{0.85}{
    \begin{tabular}{lccccccc}
        \toprule
        \multirow{2}{*}{Method} & \multirow{2}{*}{Training} & \multirow{2}{*}{\makecell{mAP\\(\%)}} & \multirow{2}{*}{\makecell{GPU Mem\\(GPUs $\times$ GB)}} & \multirow{2}{*}{\makecell{Time\\(min/epoch)}} & \multirow{2}{*}{\makecell{Param\\(M)}} \\
        & & &\\
        \midrule 
        \multirow{3}{*}{LSTR} 
        & Feat.(S) & 51.6 & $1\times1.8$ & 1.5 & 19.8 \\
        & Feat.(L+S) & $56.8$ & $1\times 2.9$  & 3.4 & 58.0 \\ 
        & E2E (L+S) & $59.2$ & $8\times 31.4$ & 7.0 & 105.9 \\ 
        \midrule 
        \multirow{2}{*}{E2E-LOAD} 
        & E2E (S) & $71.5$ & $8\times15.3$ & 6.5 & 34.2 \\ 
        & E2E (L+S) & $72.4$ & $8\times 16.9$ & 9.6 & 53.5 \\ 
        \bottomrule 
    \end{tabular}} 
    \caption{\textbf{Comparison of Training Costs.} ``S'' and ``L'' denote short-term and long-term history, respectively. ``GPU Mem'' represents the consumption of GPU memory.}
    \label{tab:training_cost}
\end{table} 

\subsubsection{Choice of Efficient Attention.}
We explore various efficient attention mechanisms proposed by the previous video models, such as the Video Swin Transformer~\cite{liu2022video}, MeMViT~\cite{wu2022memvit}, and MViT~\cite{fan2021multiscale}. Specifically, Video Swin Transformer incorporates a unique method known as shifted window attention, which decomposes the video clip into smaller windows and performs attention calculations across different hierarchical levels. On the other hand, both MeMViT and MViT employ pooling attention techniques to craft multi-scale representations. While these two approaches share similarities, the ordering of the linear layer and the pooling operation differs, leading to subtle variations in computational complexity. When comparing the performance, we did not introduce the long-term history to simplify the problem. 
Table~\ref{tab:exp:ea} details the comparative analysis among these techniques. Considering performance and training costs, we adopt pooling attention from MViT.
\begin{table}[t] 
    \centering
    \footnotesize 
    \scalebox{0.9}{
    \begin{tabular}{cccccc}
        \toprule
        \multirow{2}{*}{Method}  & \multirow{2}{*}{mAP (\%)} & \multirow{2}{*}{\makecell{GPU Mem\\(GPUs$\times$ GB)}} & \multirow{2}{*}{\makecell{Time\\(min/epoch)}} \\
        & & \\
        \midrule 
        Video Swin & $64.7$ & $8\times 12.7$ & $4.3$ \\ 
        MeMViT & $70.9$ & $8\times 14.9$  & $5.8$ \\ 
        Ours~(MViT) & $71.5$ & $8\times 15.3$ & $6.5$ \\ %
        \bottomrule 
    \end{tabular}} 
    \caption{\textbf{Comparison of Different Efficient Attention.}\label{tab:exp:ea}}
\end{table}
\subsubsection{Impact of Spatiotemporal Exploration.}
\label{sec:exp:sm_tf}
Increasing the number of spatiotemporal attention layers will lead to more computational costs while resulting in effective representations. We conducted several experiments based on \emph{Baseline+EI} to investigate the trade-off between effectiveness and efficiency.
Specifically, we control the proportion of spatiotemporal attention by adjusting the number of layers in SB and SM while maintaining the total number of layers of SB and SM.
As shown in Figure~\ref{fig:exp:sm_depth}, with the increasing $L_{SM}$, the model's performance is significantly improved, but the corresponding inference time also increases. To ensure both effectiveness and efficiency, we choose the setting of $L_{SB}=5, L_{SM}=11$. 
\subsubsection{Design of the Short-term Modeling.} 
For the SM module, we conduct experiments based on \emph{Baseline+EI} to investigate the impact of the short-term history's length. Shown in Figure~\ref{fig:exp:short_len}, as the period $T_S$ increases, the performance gradually increases, and the FPS gradually decreases, which indicates that a wider receptive field will provide more spatiotemporal clues for the ongoing actions, accompanied by a more considerable computational burden due to the element-wise interactions. We take $T_S=32$ for the trade-off of effectiveness and efficiency. 
\subsubsection{Design of the Long-term Compression.}
The LC module encodes the long-term histories as compact representations to enrich the short-term context. We conduct extensive experiments to study the temporal compression factor at different layers and the long-term history length $T_{L}$. All the studies are conducted on \emph{Baseline+LC+LSF+EI}. From Table~\ref{tab:exp:lc}, the $\times 2, \times 2, \times 1, \times 1$ outperforms the other settings, indicating the importance of the progressive compression. 
From Figure~\ref{fig:exp:long_len}, when the length of long-term history $T_{L}$ is 32, it is the most helpful for training, compared with other settings.
We set the compression factor to $\times 2, \times 2, \times 1, \times 1$ and the long-term historical length to $32$ for training.
\begin{figure}[t]
\centering
\includegraphics[width=8.0cm]{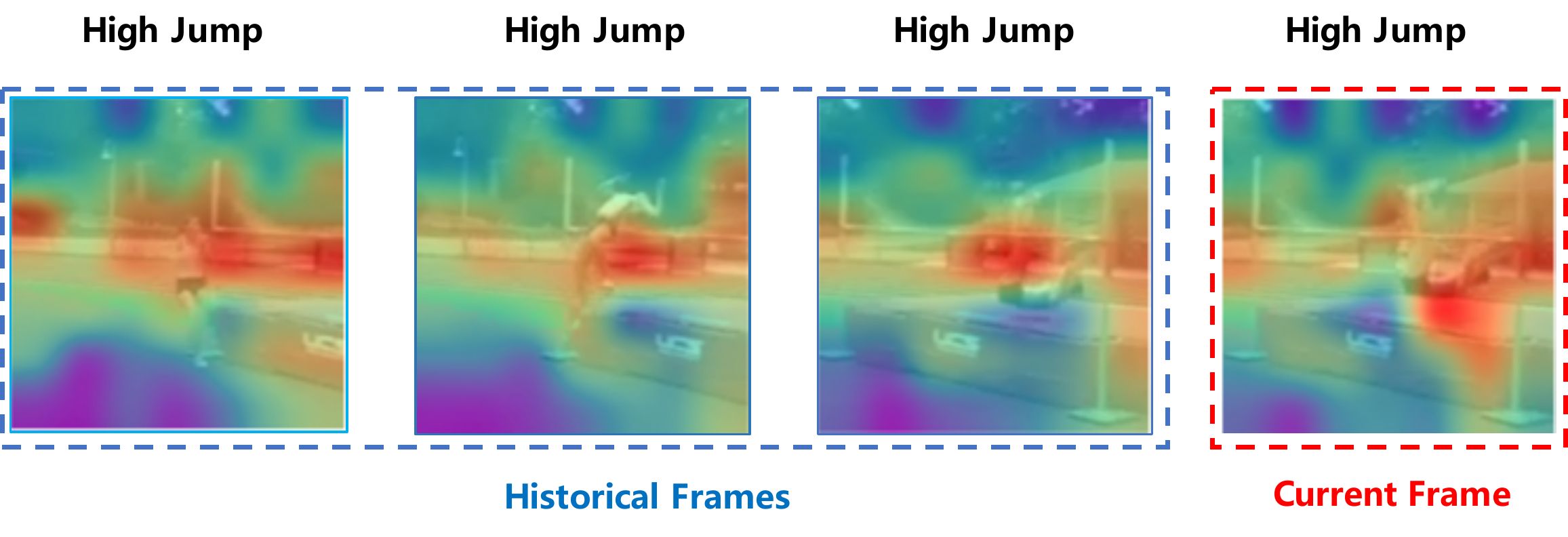}
\caption{The visualization of the spatiotemporal attention in the SM module. It illustrates the attention distributions of the current frame~(red dotted box) on the historical frames~(blue dotted box). }
\label{fig:vis}
\end{figure}
\subsubsection{Design of the Long-Short-term Fusion.}
The LSF module is designed to fuse the compressed long-term history with the short-term history. We study the validation of the fusion operation and the fusion position based on \emph{Baseline+LC+LSF+EI}. We first discuss which layer of the SM module to fuse the compressed history is the best option, shown in Table~\ref{tab:exp:fusion_pos}. We define three fusion types, \ie early fusion, middle fusion, and late fusion, and observe that the middle fusion will result in the best performance. 
This is because early fusion may cause a misalignment of features since the compressed historical tokens are well explored at the spatiotemporal dimension. In contrast, the short-term historical tokens are not well characterized at early stages. As for the late fusion, we observe over-fitting in earlier iterations. This is because the Long-term Compression~(LC) module contains fewer parameters than the Short-term Modeling~(SM), leading to over-fitting and dominating the fusion. So we employ fusion at the $5^{th}$ layer for our E2E-LOAD. Besides, we discuss the impact of fusion operations, i.e., cross-attention~(CA) or self-attention~(SA), as shown in Table~\ref{tab:exp:fusion_ops}. We adopt the cross-attention as the fusion operations due to the superior performance. 
\subsubsection{Generalization of Sequence Length.}
End-to-end training with long-term historical sequences is challenging due to the enormous resource consumption. To address this issue, we intend to investigate the ability of the Transformer models to generalize to longer sequences. This allows us to use relatively short histories during the training process and sufficiently long histories during the test process for the LC module. As shown in Figure~\ref{fig:generalization}, we observe that with or without EI, E2E-LOAD produces better performance for longer sequences during inference while training is limited to 32 frames. 
Therefore, during the inference process, we extend the long-term historical frame to 128, and further extension beyond this length does not yield significant performance improvement.
\subsection{Running Time}
As shown in Figure~\ref{fig:teaser} and Table~\ref{tab:sota}, we compare E2E-LOAD with other approaches in terms of running time, which is tested on Tesla V100. We can observe that feature-based methods are constrained by optical flow extraction and infer at around 5 FPS. Once the optical flow is removed, although the speed can be significantly improved, the performance is also considerably frustrating. Instead, our E2E-LOAD can efficiently run at 8.7 FPS. With EI, it can run at 17.3 FPS while retaining performance.
\subsection{Visualization}
We qualitatively validate the effectiveness of E2E-LOAD by visualizing a spatiotemporal attention map in the current window. Figure~\ref{fig:vis} shows a ``High Jump`` demo where we observe a strong correlation of the subject in the current and historical frames, and the irrelevant background can be well suppressed. More examples can be found in the supplementary material.
\section{Conclusion}
\label{sec:conclusion}
This paper proposes E2E-LOAD, an end-to-end framework based on Transformers for online action detection. Our framework addresses the critical challenges of OAD, including long-term understanding and efficient inference, with novel designs such as stream buffer, short-term modeling, long-term compression, long-short-term fusion, and efficient inference. Through extensive experiments on three benchmarks, E2E-LOAD achieves higher efficiency and effectiveness than existing approaches as E2E-LOAD provides an efficient framework for modeling long videos, which may be helpful for other long-form video tasks.
\section*{Acknowledgement} This work was supported in part by the National Key Research and Development Plan of China under Grant 2021ZD0112002, in part by the National Natural Science Foundation of China under Grant U1913204 and Grant 61991411, in part by the Natural Science Foundation of Shandong Province for Distinguished Young Scholars under Grant ZR2020JQ29, and in part by Project for Self-Developed Innovation Team of Jinan City under Grant 2021GXRC038.
{\small
\bibliographystyle{ieee_fullname}
\bibliography{egbib}

\begin{thebibliography}{10}\itemsep=-1pt

\bibitem{arnab2021vivit}
Anurag Arnab, Mostafa Dehghani, Georg Heigold, Chen Sun, Mario Lu{\v{c}}i{\'c},
  and Cordelia Schmid.
\newblock Vivit: A video vision transformer.
\newblock In {\em Proceedings of the IEEE/CVF International Conference on
  Computer Vision}, pages 6836--6846, 2021.

\bibitem{bertasius2021space}
Gedas Bertasius, Heng Wang, and Lorenzo Torresani.
\newblock Is space-time attention all you need for video understanding?
\newblock In {\em ICML}, volume~2, page~4, 2021.

\bibitem{chen2022gatehub}
Junwen Chen, Gaurav Mittal, Ye Yu, Yu Kong, and Mei Chen.
\newblock Gatehub: Gated history unit with background suppression for online
  action detection.
\newblock In {\em Proceedings of the IEEE/CVF Conference on Computer Vision and
  Pattern Recognition}, pages 19925--19934, 2022.

\bibitem{cheng2022stochastic}
Feng Cheng, Mingze Xu, Yuanjun Xiong, Hao Chen, Xinyu Li, Wei Li, and Wei Xia.
\newblock Stochastic backpropagation: a memory efficient strategy for training
  video models.
\newblock In {\em Proceedings of the IEEE/CVF Conference on Computer Vision and
  Pattern Recognition}, pages 8301--8310, 2022.

\bibitem{eun2020learning}
Hyunjun Eun, Jinyoung Moon, Jongyoul Park, Chanho Jung, and Changick Kim.
\newblock Learning to discriminate information for online action detection.
\newblock In {\em Proceedings of the IEEE/CVF Conference on Computer Vision and
  Pattern Recognition}, pages 809--818, 2020.

\bibitem{fan2021multiscale}
Haoqi Fan, Bo Xiong, Karttikeya Mangalam, Yanghao Li, Zhicheng Yan, Jitendra
  Malik, and Christoph Feichtenhofer.
\newblock Multiscale vision transformers.
\newblock In {\em Proceedings of the IEEE/CVF International Conference on
  Computer Vision}, pages 6824--6835, 2021.

\bibitem{feichtenhofer2019slowfast}
Christoph Feichtenhofer, Haoqi Fan, Jitendra Malik, and Kaiming He.
\newblock Slowfast networks for video recognition.
\newblock In {\em Proceedings of the IEEE/CVF international conference on
  computer vision}, pages 6202--6211, 2019.

\bibitem{gao2019startnet}
Mingfei Gao, Mingze Xu, Larry~S Davis, Richard Socher, and Caiming Xiong.
\newblock Startnet: Online detection of action start in untrimmed videos.
\newblock In {\em Proceedings of the IEEE/CVF International Conference on
  Computer Vision}, pages 5542--5551, 2019.

\bibitem{gao2021woad}
Mingfei Gao, Yingbo Zhou, Ran Xu, Richard Socher, and Caiming Xiong.
\newblock Woad: Weakly supervised online action detection in untrimmed videos.
\newblock In {\em Proceedings of the IEEE/CVF Conference on Computer Vision and
  Pattern Recognition}, pages 1915--1923, 2021.

\bibitem{geest2016online}
Roeland~De Geest, Efstratios Gavves, Amir Ghodrati, Zhenyang Li, Cees Snoek,
  and Tinne Tuytelaars.
\newblock Online action detection.
\newblock In {\em European Conference on Computer Vision}, pages 269--284.
  Springer, 2016.

\bibitem{guouncertainty}
Hongji Guo, Zhou Ren, Yi Wu, Gang Hua, and Qiang Ji.
\newblock Uncertainty-based spatial-temporal attention for online action
  detection.
\newblock 2022.

\bibitem{hochreiter1997long}
Sepp Hochreiter and J{\"u}rgen Schmidhuber.
\newblock Long short-term memory.
\newblock {\em Neural computation}, 9(8):1735--1780, 1997.

\bibitem{idrees2017thumos}
Haroon Idrees, Amir~R Zamir, Yu-Gang Jiang, Alex Gorban, Ivan Laptev, Rahul
  Sukthankar, and Mubarak Shah.
\newblock The thumos challenge on action recognition for videos “in the
  wild”.
\newblock {\em Computer Vision and Image Understanding}, 155:1--23, 2017.

\bibitem{YoungHwiKim2021TemporallySO}
Young~Hwi Kim, Seonghyeon Nam, and Seon~Joo Kim.
\newblock Temporally smooth online action detection using cycle-consistent
  future anticipation.
\newblock {\em Pattern Recognition}, 2021.

\bibitem{li2022mvitv2}
Yanghao Li, Chao-Yuan Wu, Haoqi Fan, Karttikeya Mangalam, Bo Xiong, Jitendra
  Malik, and Christoph Feichtenhofer.
\newblock Mvitv2: Improved multiscale vision transformers for classification
  and detection.
\newblock In {\em Proceedings of the IEEE/CVF Conference on Computer Vision and
  Pattern Recognition}, pages 4804--4814, 2022.

\bibitem{liu2021swin}
Ze Liu, Yutong Lin, Yue Cao, Han Hu, Yixuan Wei, Zheng Zhang, Stephen Lin, and
  Baining Guo.
\newblock Swin transformer: Hierarchical vision transformer using shifted
  windows.
\newblock In {\em Proceedings of the IEEE/CVF International Conference on
  Computer Vision}, pages 10012--10022, 2021.

\bibitem{liu2022video}
Ze Liu, Jia Ning, Yue Cao, Yixuan Wei, Zheng Zhang, Stephen Lin, and Han Hu.
\newblock Video swin transformer.
\newblock In {\em Proceedings of the IEEE/CVF Conference on Computer Vision and
  Pattern Recognition}, pages 3202--3211, 2022.

\bibitem{neimark2021video}
Daniel Neimark, Omri Bar, Maya Zohar, and Dotan Asselmann.
\newblock Video transformer network.
\newblock In {\em Proceedings of the IEEE/CVF International Conference on
  Computer Vision}, pages 3163--3172, 2021.

\bibitem{qu2020lap}
Sanqing Qu, Guang Chen, Dan Xu, Jinhu Dong, Fan Lu, and Alois Knoll.
\newblock Lap-net: Adaptive features sampling via learning action progression
  for online action detection.
\newblock {\em arXiv preprint arXiv:2011.07915}, 2020.

\bibitem{ramanishka2018CVPR}
Vasili Ramanishka, Yi-Ting Chen, Teruhisa Misu, and Kate Saenko.
\newblock Toward driving scene understanding: A dataset for learning driver
  behavior and causal reasoning.
\newblock In {\em Conference on Computer Vision and Pattern Recognition
  (CVPR)}, 2018.

\bibitem{shou2017cdc}
Zheng Shou, Jonathan Chan, Alireza Zareian, Kazuyuki Miyazawa, and Shih-Fu
  Chang.
\newblock Cdc: Convolutional-de-convolutional networks for precise temporal
  action localization in untrimmed videos.
\newblock In {\em Proceedings of the IEEE conference on computer vision and
  pattern recognition}, pages 5734--5743, 2017.

\bibitem{shou2018online}
Zheng Shou, Junting Pan, Jonathan Chan, Kazuyuki Miyazawa, Hassan Mansour,
  Anthony Vetro, Xavier Giro-i Nieto, and Shih-Fu Chang.
\newblock Online detection of action start in untrimmed, streaming videos.
\newblock In {\em Proceedings of the European Conference on Computer Vision
  (ECCV)}, pages 534--551, 2018.

\bibitem{simonyan2014two}
Karen Simonyan and Andrew Zisserman.
\newblock Two-stream convolutional networks for action recognition in videos.
\newblock {\em Advances in neural information processing systems}, 27, 2014.

\bibitem{vaswani2017attention}
Ashish Vaswani, Noam Shazeer, Niki Parmar, Jakob Uszkoreit, Llion Jones,
  Aidan~N Gomez, {\L}ukasz Kaiser, and Illia Polosukhin.
\newblock Attention is all you need.
\newblock {\em Advances in neural information processing systems}, 30, 2017.

\bibitem{wang2016temporal}
Limin Wang, Yuanjun Xiong, Zhe Wang, Yu Qiao, Dahua Lin, Xiaoou Tang, and Luc
  Van~Gool.
\newblock Temporal segment networks: Towards good practices for deep action
  recognition.
\newblock In {\em European conference on computer vision}, pages 20--36.
  Springer, 2016.

\bibitem{wang2021oadtr}
Xiang Wang, Shiwei Zhang, Zhiwu Qing, Yuanjie Shao, Zhengrong Zuo, Changxin
  Gao, and Nong Sang.
\newblock Oadtr: Online action detection with transformers.
\newblock In {\em Proceedings of the IEEE/CVF International Conference on
  Computer Vision}, pages 7565--7575, 2021.

\bibitem{wu2022memvit}
Chao-Yuan Wu, Yanghao Li, Karttikeya Mangalam, Haoqi Fan, Bo Xiong, Jitendra
  Malik, and Christoph Feichtenhofer.
\newblock Memvit: Memory-augmented multiscale vision transformer for efficient
  long-term video recognition.
\newblock In {\em Proceedings of the IEEE/CVF Conference on Computer Vision and
  Pattern Recognition}, pages 13587--13597, 2022.

\bibitem{xu2019temporal}
Mingze Xu, Mingfei Gao, Yi-Ting Chen, Larry~S Davis, and David~J Crandall.
\newblock Temporal recurrent networks for online action detection.
\newblock In {\em Proceedings of the IEEE/CVF International Conference on
  Computer Vision}, pages 5532--5541, 2019.

\bibitem{xu2021long}
Mingze Xu, Yuanjun Xiong, Hao Chen, Xinyu Li, Wei Xia, Zhuowen Tu, and Stefano
  Soatto.
\newblock Long short-term transformer for online action detection.
\newblock {\em Advances in Neural Information Processing Systems}, 34, 2021.

\bibitem{yang2022colar}
Le Yang, Junwei Han, and Dingwen Zhang.
\newblock Colar: Effective and efficient online action detection by consulting
  exemplars.
\newblock In {\em Proceedings of the IEEE/CVF Conference on Computer Vision and
  Pattern Recognition}, pages 3160--3169, 2022.

\bibitem{zhao2020privileged}
Peisen Zhao, Lingxi Xie, Ya Zhang, Yanfeng Wang, and Qi Tian.
\newblock Privileged knowledge distillation for online action detection.
\newblock {\em arXiv preprint arXiv:2011.09158}, 2020.

\bibitem{zhao2022real}
Yue Zhao and Philipp Kr{\"a}henb{\"u}hl.
\newblock Real-time online video detection with temporal smoothing
  transformers.
\newblock {\em arXiv preprint arXiv:2209.09236}, 2022.

\bibitem{zhu2020comprehensive}
Yi Zhu, Xinyu Li, Chunhui Liu, Mohammadreza Zolfaghari, Yuanjun Xiong, Chongruo
  Wu, Zhi Zhang, Joseph Tighe, R Manmatha, and Mu Li.
\newblock A comprehensive study of deep video action recognition.
\newblock {\em arXiv preprint arXiv:2012.06567}, 2020.

\end{thebibliography}
}

\end{document}